\documentclass{article}

\usepackage{arxiv}

\usepackage[utf8]{inputenc} 
\usepackage[T1]{fontenc}    
\usepackage{hyperref}       
\usepackage{url}            
\usepackage{booktabs}       
\usepackage{amsfonts}       
\usepackage{nicefrac}       
\usepackage{microtype}      
\usepackage{graphicx}
\usepackage{natbib}
\usepackage{doi}

\title{Evaluation of Environmental Conditions on Object Detection \\
using Oriented Bounding Boxes for AR Applications}

\author{ \href{https://orcid.org/0000-0003-0298-4931}{\includegraphics[scale=0.06]{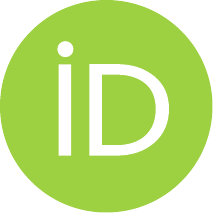}\hspace{1mm}Vladislav Li} \\
	Department of Networks and Digital Media\\
	Kingston University\\
	London, UK \\
	\texttt{v.li@kingston.ac.uk} \\
	\And
	\href{https://orcid.org/0000-0002-2846-0610}{\includegraphics[scale=0.06]{orcid.pdf}\hspace{1mm}Barbara Villarini} \\
	School of Computer Science and Engineering\\
	University of Westminster\\
	London, UK \\
	\texttt{b.villarini@westminster.ac.uk} \\
 	\And
	\href{https://orcid.org/0000-0003-1812-5269}{\includegraphics[scale=0.06]{orcid.pdf}\hspace{1mm}Jean-Christophe Nebel} \\
	Department of Computer Science\\
	Kingston University\\
	London, UK \\
	\texttt{j.nebel@kingston.ac.uk} \\
 	\And
	\href{https://orcid.org/0000-0002-0749-9794}{\includegraphics[scale=0.06]{orcid.pdf}\hspace{1mm}Thomas Lagkas} \\
	Department of Computer Science\\
	International Hellenic University\\
	Greece \& S.E. Research Centre \\
	\texttt{tlagkas@ieee.org} \\
 	\And
	\href{https://orcid.org/0000-0001-6042-0355}{\includegraphics[scale=0.06]{orcid.pdf}\hspace{1mm}Panagiotis Sarigiannidis} \\
	Department of Electrical and Comp. Engineering\\
	University of Western Macedonia\\
	Kozani, Greece \\
	\texttt{psarigiannidis@uowm.gr} \\
 	\And
	\href{https://orcid.org/0000-0003-4679-8049}{\includegraphics[scale=0.06]{orcid.pdf}\hspace{1mm}Vasileios Argyriou} \\
	Department of Networks and Digital Media\\
	Kingston University\\
	London, UK \\
	\texttt{vasileios.argyriou@kingston.ac.uk} \\
}

\date{}

\hypersetup{
pdftitle={Evaluation of Environmental Conditions on Object Detection \\
using Oriented Bounding Boxes for AR Applications},
pdfsubject={cs.AI, cs.AI},
pdfauthor={Vladislav Li, Barbara Villarini, Jean-Christophe Nebel, Thomas Lagkas, Panagiotis Sarigiannidis, Vasileios Argyriou},
pdfkeywords={artificial intelligence, augmented reality, oriented-bounding boxes, synthetic dataset},
}

\begin{document}
\maketitle

\begin{abstract}
The objective of augmented reality (AR) is to add digital content to natural images and videos to create an interactive experience between the user and the environment. Scene analysis and object recognition play a crucial role in AR, as they must be performed quickly and accurately. In this study, a new approach is proposed that involves using oriented bounding boxes with a detection and recognition deep network to improve performance and processing time. The approach is evaluated using two datasets: a real image dataset (DOTA dataset) commonly used for computer vision tasks, and a synthetic dataset that simulates different environmental, lighting, and acquisition conditions. The focus of the evaluation is on small objects, which are difficult to detect and recognise. The results indicate that the proposed approach tends to produce better Average Precision and greater accuracy for small objects in most of the tested conditions.
\end{abstract}

\keywords{artificial intelligence \and augmented reality \and oriented-bounding boxes \and synthetic dataset}

\section{Introduction}
By projecting digital material, such as text, images, videos, or 3D models \cite{ARGYRIOU2010887} over actual environments in real-time utilising the camera view, augmented reality (AR) apps enable users to engage with their surroundings. \cite{tianyu2017overview} The experience of the real world is so improved. The need for human-digital interaction via Mixed Reality (XR) headsets and new 3D interactive displays has surged as a result of developments in computer systems, high-speed connectivity, and computer vision technologies. The quick advancement of AR technology has made it possible for it to be used in a variety of fields, including restoration, education, behaviour and motion analysis \cite{Argyriou2011SubHexagonalPC,10111712762230,ip-vis_20051073,6977392}, archaeology, art, tourism, commerce, healthcare, etc. \cite{xiong2021augmented,8613651}

It is essential for immersive technologies to analyse the environment around them in order to extract pertinent information. Scene analysis and comprehension, for instance, are crucial for autonomous vehicle decision-making and end-to-end vision control \cite{pavel2022vision}, comprising tasks like vehicle identification, traffic sign and light recognition, and pedestrian detection. Once this data has been rendered, the augmented environment can be seen visually on the vehicle's display. The process of extracting this information may be improved by the use of oriented bounding boxes. Furthermore, the scene analysis would benefit from the synthetic dataset created to represent particular environmental conditions.

The development of more effective object identification techniques as a result of recent developments in computer vision has increased computational efficiency and decreased processing time \cite{zou2019jieping}, which is crucial for the object detection task \cite{sermanet2013overfeat}. Complex computer vision algorithms are used by object recognition-based AR technology to recognise and track items in the real environment. These include the You Only Look Once (YOLO) model \cite{anderson2019feasibility}, homomorphic filtering and Haar markers \cite{gomes2018augmented}, and the Single Shot Detector \cite{dimitropoulos2021operator}. Faster and more precise object recognition has been achieved by the combination of Convolutional Neural Networks (CNNs) and Deep Learning (DL) \cite{zhao2019object}. The object detection process can still face difficulties due to small or distant objects or insufficient camera quality, which can also have an impact on the overall AR experience.

The goal of this work was to suggest a novel method that improves item recognition and prediction by using oriented bounding boxes instead of traditional axis-aligned bounding boxes. The network's YOLO series architecture was used in its construction with the goal of enhancing performance while lowering computational expense. The study's main accomplishments included: a) a method to deal with the problem of strong deviation angle loss and a quicker method for multi-scale feature fusion; b) a thorough comparison of performance under different environmental conditions that affect AR devices; and c) the development of a new dataset containing synthetic objects under various conditions, which was used to evaluate performance impartially across various sensor and environmental parameters.

When the suggested method is compared to the standard two-stage and single-stage procedures, the results consistently show a positive trend, suggesting improved performance. Furthermore, during a more in-depth fine-tuning process, the model produced even better results across the majority of the experiments performed. This shows that the new method has the ability to successfully solve the shortcomings of prior approaches while also providing more accurate and reliable item identification and recognition in a variety of applications. The model's capacity to adapt to varied settings and tackle complicated tasks accounts for the improved performance, making it a potential alternative for future research and real-world applications.

The paper is structured as follows: In section 1, the problem and relevant technologies are introduced. Section 2 provides a review of related research. The proposed end-to-end architecture is detailed in section 3. Section 4 showcases the results using both a real image dataset (DOTA) and a new synthetic image dataset. An ablation study is described in section 5. Finally, the conclusion is presented in section 6.

\section{Literature Review}

An AR app identifies objects in the real world using machine learning and computer vision techniques with the goal of overlaying virtual objects in real-time. In recent years, the use of deep CNNs \cite{lecun1998gradient} has greatly enhanced the performance and accuracy of object detection and recognition in computer vision. In 2014, Girishick et al. introduced the Regions with CNN features (RCNN) method for object detection \cite{girshick2014rich}. This approach involved first identifying potential object boxes through selective search and then rescaling each box to a fixed-size image for input into a CNN model trained on AlexNet \cite{viola2001rapid} for feature extraction. The object was then detected using a linear SVM classifier, resulting in a significant improvement in mean Average Precision compared to previous methods, but also had a significant drawback of slow detection speed.

He et. al. introduced the Spatial Pyramid Pooling Network (SPPNet) to resolve the slow detection speed problem \cite{he2015spatial}. The network uses a Spatial Pyramid Pooling (SPP) layer to generate a fixed-length representation that is invariant to image size and scale, which reduces overfitting and allows the network to handle images of varying sizes during training. With object detection, the feature maps are calculated only once for the entire image, and then the features in sub-regions are pooled to generate fixed-length representations for detector training. During testing, this method is 24-102 times faster than the RCNN approach.

In 2015, Girishick improved the previous two architectures with Fast RCNN \cite{girshick2015fast}. This network trains both a detector and a bounding box regression simultaneously with the same configuration. However, the speed limitation persisted. The same year, Ren et al. introduced the Faster RCNN detector \cite{ren2015faster}, which was the first deep learning detector to almost achieve real-time detection through end-to-end training. This architecture employed the Region Proposal Network (RPN) to speed up the detection process, and several variants have been proposed since then to reduce computational redundancy \cite{dai2016r}, \cite{li2017light}, \cite{lin2017feature}. In particular, Cao et al. (2020) \cite{cao2020d2det} introduced the D2Det method based on the Faster R-CNN framework, which processes Region of Interest (ROI) features through two stages: high-density local regression and discriminant ROI pooling. The method replaces the Faster RCNN offset regression with a local dense regression block.

\begin{figure}
    \centering
    \includegraphics[scale=0.50]{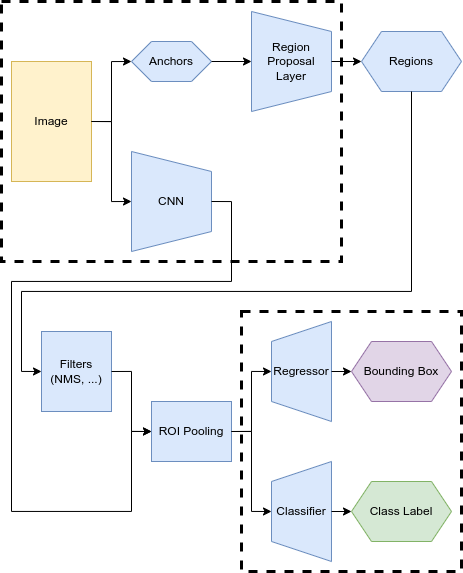}
    \caption{An abstract of the FRRCNN architecture.}
    \label{fig:abstract-frrcnn-architecture}
\end{figure}

The above methods were classified as two-stage detectors because of their two-step procedure: first generating regions of interest (ROIs) and then performing detection and recognition. In 2016, Joseph et al. proposed a one-stage detector named You Only Look Once (YOLO) \cite{redmon2016you}. This method uses a single network that processes the entire image in one step, resulting in faster processing times. The image is divided into regions and the network predicts bounding boxes for each region simultaneously. In the following years, YOLO v2 and v3 \cite{redmon2017yolo9000}, \cite{redmon2018yolov3} were proposed with the goal of improving prediction accuracy. While YOLO was faster than the two-stage methods, it resulted in lower localisation accuracy. To address this, Liu et al. introduced the Single Shot MultiBox Detector (SSD) \cite{liu2016ssd}, which utilized multi-reference and multi-resolution detection methods to detect objects at different scales across different network layers. In 2018, Lin et al. improved upon this by presenting RetinaNet \cite{lin2017focal}, which added a new loss function called "focal loss." This modification to the standard cross-entropy loss caused the detector to be more attentive to misclassified examples during training, leading to improved accuracy compared to one-stage methods.

\begin{figure}
    \centering
    \includegraphics[scale=0.40]{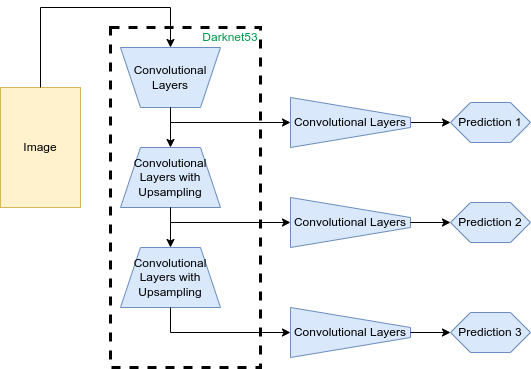}
    \caption{An abstract of the YOLOv3 architecture.}
    \label{fig:abstract-yolov3-architecture}
\end{figure}

In this study, we propose a novel method for detecting and recognising small objects in augmented reality settings. Our approach involves modifying the YOLO architecture by incorporating oriented bounding boxes. This modification enhances the accuracy of detecting objects located at large distances and at odd angles from the camera sensor of AR glasses, while maintaining low processing time for real-time application.

The DOTA dataset \cite{xia2018dota} is a collection of satellite images produced by Google Earth, GF-2, JL-1, and CycloMedia. The images represent distant objects position at random locations and arbitrary rotation. Further, the dataset is annotated in a specific format containing information about all four vertices of a bounding box making it suitable for oriented-bounding box prediction unlike more common format of a corner position accompanied by width and height. The dataset mainly consisted of RGB images with categories such as helicopter, small vehicle, large vehicle, etc.

\begin{figure}
    \centering
    \includegraphics[scale=0.20]{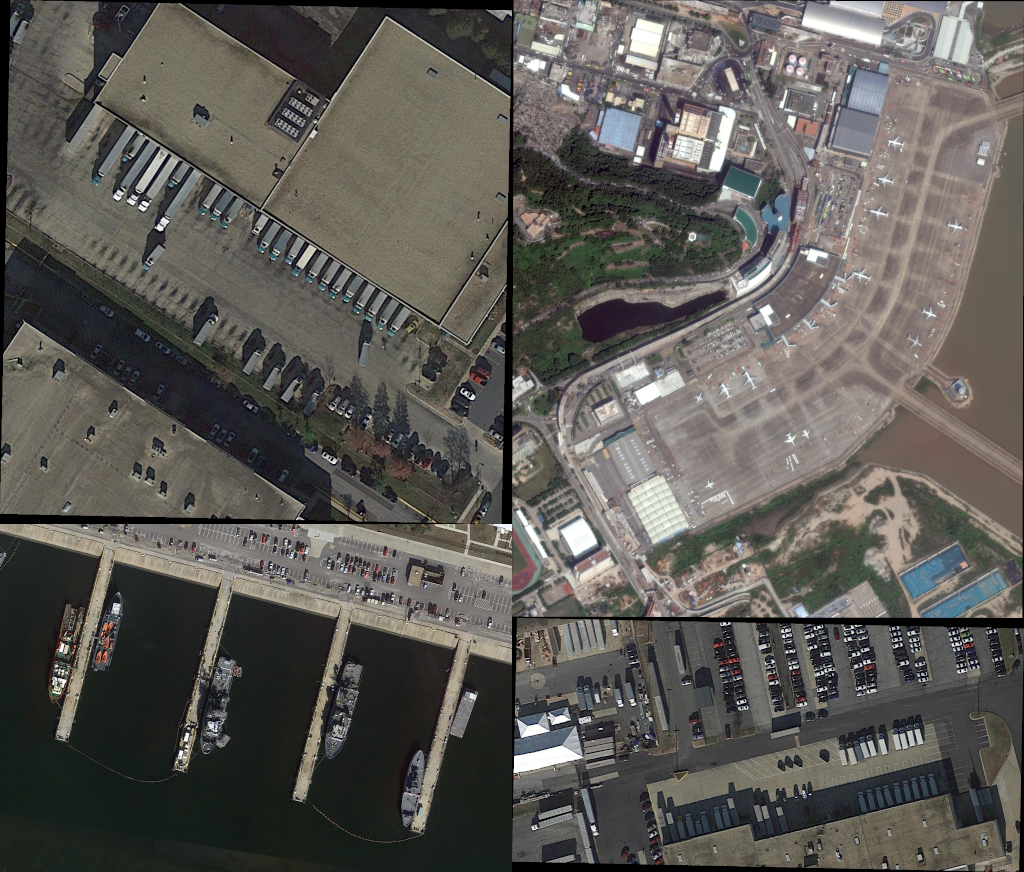}
    \caption{An abstract of the YOLOv3 architecture.}
    \label{fig:dota-example}
\end{figure}

Further assessment was carried out using the Synthetic dataset comprised of low-resolution images of various vehicles in diverse environments and weather conditions. This dataset consisted of images of different types of vehicles captured in various environments and weather conditions. The images were generated using a 3D rendering engine and organised into four categories to evaluate the model's specific properties, including camera, light, weather, and sensor. The data generation pipeline involved capturing 3D models in various 3D environments. The output would contain the images of different types with a segmentation map which is then processed to generate bounding boxes. The result is further processed using a Python script to structure the data in an appropriate format required for the AI model.

\begin{table*}[t]
    \centering
    \caption{Results of the model on the Synthetic dataset.}
    \begin{tabular}{|c|c|c|c|c|}
        \hline
        \textbf{category} & \textbf{air (10 epochs)} & \textbf{air (100 epochs)} & \textbf{ground (10 epochs)} & \textbf{ground (100 epochs)} \\
        \hline
        camera   & 28.13           & \textbf{74.35}  & 76.06                      & \textbf{88.12} \\
        light    & 82.00           & \textbf{89.25}  & 81.85                      & \textbf{82.84} \\
        sensor   & 60.73           & \textbf{71.92}  & 51.88                      & \textbf{83.66} \\
        weather  & 23.57           & \textbf{44.11}  & \textbf{26.21}             & 20.37 \\
        \hline
    \end{tabular}
    \label{table:synthetic-results}
\end{table*}

\begin{table*}[t]
    \centering
    \caption{Results of the model on the Real dataset.}
    \begin{tabular}{|c|c|c|}
        \hline
        \textbf{category} & \textbf{10 epochs} & \textbf{100 epochs} \\
        \hline
        real              & 70.28              & \textbf{76.51}      \\
        \hline
    \end{tabular}
    \label{table:real-results}
\end{table*}

\section{Methodology}

This section provides an in-depth examination of the methodology behind the proposed model, which integrates the oriented bounding box feature to bolster its object detection capabilities. The conversation centers on the YOLOv5 model's architecture, accentuating crucial distinctions between YOLOv3 and YOLOv5. These differences include the adoption of CSPDarknet53 as the backbone, the enhanced neck utilizing the Path Aggregation Network (PANet), and the inclusion of the oriented bounding boxes module to boost the model's overall performance.

\begin{figure}
    \centering
    \includegraphics[scale=0.50]{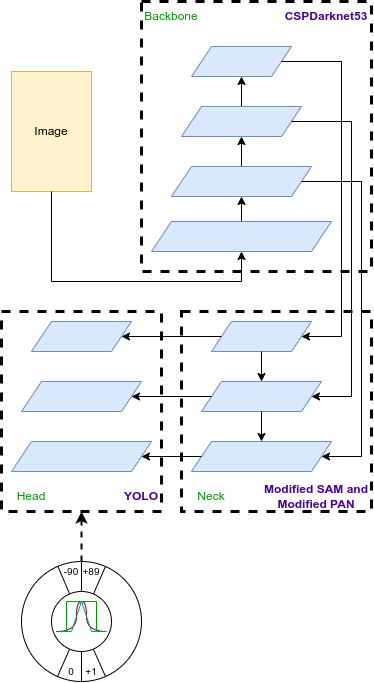}
    \caption{An abstract of the YOLOv5 with Oriented Bounding Boxes (OBB) architecture.}
    \label{fig:abstract-yolov5-architecture}
\end{figure}

The YOLO (You Only Look Once) models are a type of computer vision model that are designed to detect and classify objects within an image. They are built upon a custom backbone architecture that is based on GoogLeNet, a popular convolutional neural network (CNN) architecture. YOLO proposed its own backbone network as a faster alternative to the VGG-16 classifier. While VGG-16 is known for its accuracy in object classification, it requires a significant amount of computational power. Specifically, VGG-16 requires 30.69 billion floating point operations, making it slower when compared to YOLO classifier. The YOLO alternative is Darknet-19, which requires only 8.52 billion operations. Despite its lower computational cost, Darknet-19 is still able to achieve relatively high accuracy. Another important aspect, the YOLO architecture was designed to infer information straight from pixels to features in contrast to its competitor which utilised the sliding window technique for the object detection task.

Darknet-53 is an improvement upon Darknet-19 that integrates residual connections. Residual connections are a type of shortcut that allow information to bypass certain layers of a neural network. By integrating these connections, Darknet-53 is able to improve its performance on complex object detection tasks. Overall, YOLO models are a promising approach to object detection that balance accuracy and speed, and their custom backbone architecture allows for flexibility in adapting to different use cases. The advanced version of the single-stage object detection model, You Only Look Once (YOLO), employs a more sophisticated backbone known as CSPDarknet53. This innovative backbone is built on the foundation of the CSPNet strategy, which works by partitioning the feature map of the base layer into separate components. Following this division, the parts are combined through a cross-stage hierarchical process, enabling more efficient feature extraction.

In addition to the improved backbone, YOLO incorporates the Path Aggregation Network (PANet) as its "neck." In our case, the "neck" is an intermediate component connecting the backbone and the detection head. While the backbone extracts feature maps from input images, the detection head carries out object detection and classification. The neck acts as a bridge, aggregating and fusing features from the backbone before forwarding them to the detection head. In the context of YOLO, the Path Aggregation Network (PANet) functions as the model's "neck".

This network serves as a Feature Pyramid Network (FPN) containing a series of bottom-up and top-down layers that contribute to more effective object detection. The PANet's primary function is to aggregate and fuse features at various scales, which significantly enhances the model's overall performance. The architecture of this enhanced YOLO model establishes a streamlined pipeline where CSPDarknet53 is responsible for extracting feature maps, PANet performs feature fusion across multiple scales, and the final layer typically handles the prediction. This pipeline arrangement ensures a more efficient and accurate object detection process. To further boost performance and adapt the model to detect distant objects more effectively, an oriented bounding box module has been integrated. This specialized module allows the model to estimate the orientation of objects in the scene, providing additional context and improving detection accuracy for objects situated far away or partially occluded.

In regression-based methods for object detection, various representations of oriented bounding boxes can be employed. One such representation uses a $(x, y, w, h, \theta)$ format, where $x, y$ denote the center of a prediction, $(w, h)$ represent the width and height, and $(\theta)$ is an angle that falls within the range $0 \leq (\theta) \leq 90$. Another similar format also utilizes $(x, y, w, h, \theta)$, but in this case, $(\theta)$ is an angle in the range $0 \leq (\theta) \leq 180$. A more distinct representation employs an eight-parameter format, $(x_0, y_0, x_1, y_1, x_2, y_2, x_3, y_3)$. In this format, each pair of coordinates, $x_n, y_n$, corresponds to a corner of the oriented bounding box, with $n$ ranging from 0 to 3, such that $0 \leq (\theta) \leq 3$. This alternative representation offers a different way to define oriented bounding boxes in the context of regression-based object detection methods.

In the proposed architecture, the model has been modified to approach the prediction of oriented bounding boxes as a classification problem rather than employing the conventional regression methods discussed earlier. To facilitate easier integration, this mechanism has been implemented as a standalone module. By representing each angle of object rotation as a distinct class or category, the model is able to treat the problem as a classification task. However, when using the $(x, y, w, h, \theta)$ format, where $(\theta)$ falls within the range $0 \leq (\theta) \leq 90$, the categories corresponding to the edge angles (1 and 90 degrees) tend to converge their losses, which is undesirable from a detection standpoint. To address this issue and incorporate angle-aware context into the regular classification loss, the Circular Smooth Label (CSL) technique is employed. This approach provides essential information for the model, enhancing the accuracy of angle predictions and ultimately improving the overall performance of the oriented bounding box detection.

The oriented bounding box module was incorporated into the detector as an additional classification branch to improve object detection, particularly for distant objects. To effectively incorporate the context of angles into the training process, a specialized loss function called Circular Smooth Label (CSL) loss was introduced. The Circular Smooth Label loss function is designed to tackle the challenges of angle estimation during the training process. In conventional object detection models, angle estimation can be problematic due to the periodic nature of angles, where the difference between two angle values might not be accurately represented using a linear scale. The CSL loss function overcomes this issue by employing a circular representation of angles, which preserves the true angular difference between predictions and ground truth labels.

The CSL loss function not only enables better angle estimation but also helps in mitigating the problem of abrupt changes in gradient updates. The smoothness of the loss function ensures that the model receives continuous gradient updates during backpropagation, leading to a more stable and efficient learning process. This, in turn, results in improved detection accuracy for objects with varying orientations, especially those situated at a distance or partially occluded. Generally, the integration of the oriented bounding box module as an additional classification branch and the implementation of the Circular Smooth Label loss function significantly enhance the model's ability to account for object orientations. By providing a better representation of angles and ensuring smooth gradient updates, the CSL loss function contributes to a more robust and accurate object detection process.

The model was fine-tuned running 10 epochs with a batch size of 3 on a real data as well as synthetic data under four different categories in the fist run, and 100 epochs with the same batch size and under the same four different categories in the second run. For detector component, learning rate was set to 0.0032 with the SGD (Stochastic Gradient Descent) optimisation.

The model was subjected to a fine-tuning process, which involved training on both real and synthetic data to ensure robust performance. This fine-tuning process was conducted in two separate runs, each with a distinct number of epochs and under four different categories to diversify the training data and improve generalisation. In the first run, the model was trained for a relatively short duration, encompassing only 10 epochs with a batch size of 3. This initial training phase aimed to test the model with the real and synthetic data, allowing it to learn basic patterns and features across the four categories. The limited number of epochs in the first run ensured that the model would not overfit to the training data, providing a solid foundation for further fine-tuning.

For the second run, the training process was extended to 100 epochs, with the batch size maintained at 3 with regards to the hardware constraints. This longer training period allowed the model to delve deeper into the nuances of the real and synthetic data, learning more complex features and relationships across the four categories. The extended duration of the second run enabled the model to refine its predictions, thereby enhancing its overall performance and accuracy. Regarding the detector component, a learning rate of 0.0032 was chosen to strike a balance between the speed of convergence and the stability of the training process. To optimise the model's weights, the Stochastic Gradient Descent (SGD) algorithm was employed, a popular optimisation technique known for its effectiveness in deep learning applications. By using SGD, the model was able to navigate the complex optimisation landscape efficiently, ultimately converging to a solution that achieved a high degree of detection accuracy.

\begin{table*}
    \centering
    \caption{Results with and without Weather rain in the Synthetic dataset using models trained for 10 \& 100 epochs.}
    \begin{tabular}{|c|c|c|c|}
        \hline
        \textbf{Epochs} & \textbf{Subset} & \textbf{rain} & \textbf{no rain}     \\
        \hline
        10              & Air             & 48.84\%           & 58.11\%          \\
        100             & Air             & \textbf{80.72\%}  & \textbf{77.77\%} \\
        10              & Ground          & 56.28\%           & 43.73\%          \\
        100             & Ground          & \textbf{85.97\%}  & \textbf{85.33\%} \\
        \hline
    \end{tabular}
    \label{table:ablation-results-weather}
\end{table*}

\begin{table*}[t]
    \centering
    \caption{Results for four different Camera distances in the Air Synthetic dataset using models trained for  10 \& 100 epochs.}
    \begin{tabular}{|c|c|c|c|c|}
        \hline
        \textbf{Epochs} & \textbf{70m}     & \textbf{163m}    & \textbf{256m}    & \textbf{350m}   \\
        \hline
        10              & 36.82\%          & 35.98\%          & 48.39\%          & 22.81\%         \\
        100             & \textbf{86.35\%} & \textbf{81.45\%} & \textbf{81.55\%} & \textbf{81.87\%}\\
        \hline
    \end{tabular}
    \label{table:ablation-results-camera-air}
\end{table*}

\begin{table*}[t]
    \centering
    \caption{Results for four different camera distances in the Ground Synthetic dataset using models trained for  10 \& 100 epochs.}
    \begin{tabular}{|c|c|c|c|c|}
        \hline
        \textbf{Epochs} & \textbf{15m}     & \textbf{35m}     & \textbf{55m}     & \textbf{75m}     \\
        \hline
        10              & 31.36\%          & 66.26\%          & 82.19\%          & 73.84\%          \\
        100             & \textbf{97.49\%} & \textbf{92.17\%} & \textbf{87.57\%} & \textbf{83.65\%} \\
        \hline
    \end{tabular}
    \label{table:ablation-results-camera-ground}
\end{table*}

\section{Results}

In this work, a novel approach was proposed to tackle the task of object recognition and scene analysis. The effectiveness of this method was evaluated through a comparison with state-of-the-art approaches using the DOTA dataset \cite{xia2018dota} and a synthetic dataset that encompasses a range of environmental conditions affecting image quality. These images depict distant objects positioned at random locations and rotations. Additionally, the DOTA dataset is annotated in a unique format that includes information about all four vertices of a bounding box, making it ideal for oriented-bounding box prediction, unlike more common formats that only provide corner positions along with width and height. The dataset is composed primarily of RGB images and features categories such as helicopters, small vehicles, and large vehicles, among others.
The mean Average Precision was introduced in 2014 from Lin at al. [35] to describe the performance of the object detection based on a user-defined set of criteria. It was defined as the mean value of the average precision of the individual class

\[mAP=  \frac{1}{n} \sum_{k=1}^{n} AP_k\]

where AP is the Average Precision of the class k and n is the number of classes.

Table \ref{table:real-results} presents the results of the model on the Real dataset, comparing the performance after 10 epochs and 100 epochs of training. The table highlights the improvement in detection accuracy as a result of the extended training duration. For the Real dataset, the model achieved a detection accuracy of 70.28\% after 10 epochs of training. This initial result indicates that the model was able to learn basic patterns and features within the data during the first run of the fine-tuning process. However, when the training was extended to 100 epochs, the model's detection accuracy increased to 76.51\%. This improvement demonstrates the benefits of the longer training period, as the model was able to learn more complex features and relationships across the dataset. The result also suggests that the fine-tuning process was successful in enhancing the model's performance on the Real dataset.

Further evaluation had been carried out using the Synthetic dataset. This dataset consisted out of images of various vehicles in different environments and weather conditions. The dataset was generated with a 3D rendering engine. Subsets of images were selected from the main dataset obtaining four different categories organised in a way to evaluate the model on specific properties: a) Camera, b) Light, c) Weather d) Sensor. The “Camera” category represented images generated with different camera angles and distances from the objects. The “Light” category contained images generated using variable balanced lighting parameters. The “Weather” category represented images generated using different balanced weather parameters, including varying rain and wind conditions. The “Sensor” category defined different night and thermal vision. The performances of the proposed framework in terms of mAP are shown in Table 3 and Figure 7 for all the four categories.

In conclusion, the results presented in Table \ref{table:real-results} show that the model's performance on the Real dataset improved considerably with the extended training duration. The increase in detection accuracy from 70.28\% to 76.51\% highlights the effectiveness of the fine-tuning process in refining the model's predictions and overall performance.

Table \ref{table:synthetic-results} presents the results of the model on the Synthetic dataset, showcasing the performance after 10 epochs and 100 epochs of training for both air and ground categories. The table highlights the improvement in detection accuracy for most categories as a result of the extended training duration.

For the air category, the model demonstrated significant improvements in detection accuracy across all subcategories when comparing the results after 10 epochs and 100 epochs of training. The detection accuracy for the camera subcategory increased from 28.13\% to 74.35\%, for the light subcategory from 82.00\% to 89.25\%, for the sensor subcategory from 60.73\% to 71.92\%, and for the weather subcategory from 23.57\% to 44.11\%. These improvements indicate that the model was able to learn more intricate features and relationships within the Synthetic dataset during the extended training period, enhancing its overall performance.

Similarly, in the ground category, the model exhibited improved detection accuracy for three out of four subcategories after 100 epochs of training. The detection accuracy for the camera subcategory increased from 76.06\% to 88.12\%, for the light subcategory from 81.85\% to 82.84\%, and for the sensor subcategory from 51.88\% to 83.66\%. However, the weather subcategory experienced a slight decrease in detection accuracy, dropping from 26.21\% after 10 epochs to 20.37\% after 100 epochs. This decline might indicate overfitting or the presence of challenging samples in the weather subcategory.

The results presented in Table \ref{table:synthetic-results} demonstrate that the model's performance on the Synthetic dataset improved considerably for most subcategories with the extended training duration. While the weather subcategory in the ground category exhibited a decline in detection accuracy, the overall trend suggests that the fine-tuning process was successful in refining the model's predictions and enhancing its performance across the Synthetic dataset. As could be grasped from the aforementioned results, the Ground dataset didn't produce a dramatic increase in performance when comparing to the Air dataset. However, the Ground dataset had higher results in the initial training. The reason could be the original dataset used during the training which would commonly be biased towards cars.

As depicted in the table \ref{table:synthetic-results} and \ref{table:real-results}, the new method presents some performance improvement, in comparison with the previous state-of-the-art solutions. Furthermore, overall trend points out that even after further fine-tuning the model still kept improving and outperformed itself in the most of the experiments.

\section{Ablation Study}

Table \ref{table:ablation-results-weather} presents the results of an ablation study performed on the Synthetic dataset, comparing the model's performance when trained for 10 and 100 epochs under two different conditions: rain and no rain. The table illustrates the impact of weather conditions on the model's detection accuracy for both air and ground subsets.

In the air subset, the model's detection accuracy was higher in the presence of rain compared to no rain for both 10 and 100 epochs of training with an exception of the top row, Table \ref{table:ablation-results-weather}. Where after 10 epochs, the detection accuracy was 48.84\% with rain and 58.11\% without rain. However, after 100 epochs, the model exhibited improved performance and the trend has flipped, with the detection accuracy increasing to 80.72\% with rain and 77.77\% without rain. This suggests that the extended training duration allowed the model to better adapt to the challenges posed by the rain condition. Similarly, for the ground subset, the model showed higher detection accuracy with rain (56.28\%) compared to without rain (43.73\%) after 10 epochs of training. After 100 epochs, the detection accuracy improved substantially, reaching 85.97\% with rain and 85.33\% without rain. This indicates that the model was able to learn more complex features related to weather conditions and achieve better performance in the presence of rain after extended training. Moreover, the rain and no rain results seemed to reach a near parity.

Overall, Table \ref{table:ablation-results-weather} demonstrates that the model's performance on the Synthetic dataset is affected by weather conditions, particularly rain. Furthermore, absence of rain doesn't always guarantee better performance because there are cases where the model performs better with the rain condition. The results show that extending the training duration from 10 to 100 epochs led to a significant improvement in detection accuracy for both air and ground subsets under both rain and no rain conditions. This suggests that the fine-tuning process was successful in enabling the model to adapt to varying conditions, ultimately enhancing its overall performance.

Continuing to the Synthetic dataset, Table \ref{table:ablation-results-camera-air} presents the results of an ablation study performed on the Air Synthetic dataset, comparing the model's performance when trained for 10 and 100 epochs at four different camera distances: $70m$, $163m$, $256m$, and $350m$. The table illustrates the impact of camera distance on the model's detection accuracy.

After 10 epochs of training, the model's detection accuracy varied across the different camera distances. The highest detection accuracy was observed at a camera distance of $256m$ (48.39\%), while the lowest was at $350m$ (22.81\%). The detection accuracy for camera distances of $70m$ and $163m$ was 36.82\% and 35.98\%, respectively. These results indicate that the model's initial performance is affected by the camera distance, naturally, with the model struggling to achieve consistent detection accuracy across all the distances. However, after 100 epochs of training, the model exhibited a substantial improvement in detection accuracy for all camera distances. The detection accuracy increased to 86.35\% for $70m$, 81.45\% for $163m$, 81.55\% for $256m$, and 81.87\% for $350m$. This demonstrates that the extended training duration allowed the model to learn more complex features related to camera distance and achieve better performance across all tested distances. In addition, the detection performance became more stable hence more predictable.

Generally, Table \ref{table:ablation-results-camera-air} shows that the model's performance on the Air Synthetic dataset is influenced by camera distance. Extending the training duration from 10 to 100 epochs led to significant improvements in detection accuracy at all camera distances, suggesting that the fine-tuning process was successful in enabling the model to adapt to varying camera distances and enhance its overall performance.

Next, Table \ref{table:ablation-results-camera-ground} presents the results of an ablation study performed on the Ground Synthetic dataset, comparing the model's performance when trained for 10 and 100 epochs at four different camera distances: $15m$, $35m$, $55m$, and $75m$. The table illustrates the impact of camera distance on the model's detection accuracy.

After 10 epochs of training, the model's detection accuracy varied across the different camera distances. The highest detection accuracy was observed at a camera distance of $55m$ (82.19\%), while the lowest was at $15m$ (31.36\%). The detection accuracy for camera distances of $35m$ and $75m$ was 66.26\% and 73.84\%, respectively. These results indicate that the model's initial performance is affected by the camera distance, with the model showing better performance at larger distances. However, after 100 epochs of training, the model exhibited a substantial improvement in detection accuracy for all camera distances. The detection accuracy increased to 97.49\% for $15m$, 92.17\% for $35m$, 87.57\% for $55m$, and 83.65\% for $75m$. This demonstrates that the extended training duration allowed the model to learn more complex features related to camera distance and achieve better performance across all tested distances.

Table \ref{table:ablation-results-camera-ground} shows that the model's performance on the Ground Synthetic dataset is influenced by camera distance. Extending the training duration from 10 to 100 epochs led to significant improvements in detection accuracy at all camera distances, suggesting that the fine-tuning process was successful in enabling the model to adapt to varying camera distances and enhance its overall performance. However, in comparison to the Air subset, the results are less stable and follow a more "natural" trend because it is commonly expected for the model to drop performance with increasing distance.

\section{Conclusion}

This paper explores scene analysis techniques in Augmented Reality (AR) and proposes a modified YOLO architecture with oriented bounding boxes for object detection and recognition. The solution detects objects at large distances and odd angles while maintaining low processing times, suitable for real-time AR applications.

Existing methods are categorised into two-stage and single-stage detectors, with the proposed model evaluated using real and synthetic datasets. The evaluation results show that the proposed solution improves object detection and recognition, particularly for distant objects, compared to state-of-the-art approaches.

In conclusion, this work offers a comprehensive analysis of AR scene analysis and object detection methods, proposing a novel solution that addresses existing limitations and demonstrates potential for real-world AR applications.

\section{ACKNOWLEDGMENT}

This work was funded by UK Research and Innovation (UKRI) under the UK government’s Horizon Europe funding guarantee [grant number 10047653] and funded by the European Union [under EC Horizon Europe grant agreement number 101070181 (TALON)].

\bibliographystyle{unsrtnat}
\bibliography{references}

\end{document}